# Deep Double-Side Learning Ensemble Model for Few-Shot Parkinson Speech Recognition

Yongming Li*, Lang Zhou, Lingyun Qin, Yuwei Zeng, Yuchuan Liu, Yan Lei, Pin Wang, Fan Li

(School of microelectronics and commuication engineering, Chongqing University, China, 400044)

**Abstract**: Diagnosis and therapeutic effect assessment of Parkinson's disease (PD) based on voice data are very important, but its few-shot learning is challenging. Although deep learning is goode at automatic feature extraction, it suffers from few-shot learning problem. Therefore, the general effective method is first conduct feature extraction based on prior knowledge, and then carry out feature dimension reduction for subsequent classification. However, there are two major problems: 1) Structural information among speech features has not been mined and new features of higher quality have not been reconstructed. 2) Structural information between data samples has not been mined and new samples with higher quality have not been reconstructed. To solve these two problems, based on the existing Parkinson speech feature data set, a deep double-side learning ensemble model is designed in this paepr that can reconstruct speech features and samples deeply and simultaneously. As to feature reconstruction, an embedded stack group sparse auto-encoder is designed in this paper to conduct nonlinear feature transformation, so as to acquire new high-level deep features, and then the deep features are fused with original speech features by L1 regularization feature selection methods. As to speech sample reconstruction, a deep sample learning algorithm is designed in this paper based on iterative mean clustering to conduct samples transformation, so as to obtain new high-level deep samples. Finally, the bagging ensemble learning mode is adopted to fuse the deep feature learning algorithm and the deep samples learning algorithm together, thereby constructing a deep double-side learning ensemble model. At the end of this paper, two representative speech datasets of Parkinson's disease were used for validation. The experimental results show that the main innovation part of the algorithm is effective. For the two data sets, the mean accuracy of the proposed algorithm reaches 98.4% and 99.6% respectively, which are better than the state of art relevant algorithms.

## 1  Introduction

Speech data recognition is an important method for the diagnosis and therapeutic effect assessment of Parkinson's disease (PD)[1][2]. In recent years, some scholars have carried out research about speech based PD diagnosis and made some progress[3][4][5].

Speech feature learning is an important components of the recognition algorithm. The methods mainly include feature selection and feature extraction (transformation). Feature selection refers to the process of selecting effective feature subsets from the original feature set. The main methods include filter selection, warpper selection, embedded selection and integrated selection[6][7][8][9][10]. By filtrating out redundant features, problems such as "dimension disaster" and "overfitting" can be alleviated and the learning complexity can be reduced. However, its main drawback is the inability to form new high-level features. Feature extraction refers to the process of transforming or mapping the original features to obtain fewer new features. The main methods are as follows: PCA, ICA, LDA, LPP, etc. It mainly contributes to improve the generalization ability of the model but it removes or diminishes the physical significance of the features. There are two major problems in these studies: 1) Structural information between data features has not been mined, and new high-level features with higher quality can not been reconstructed; 2) Structural information between data samples has not been mined, and new high-level samples with higher quality have not been reconstructed.

In terms of speech features, deep feature learning methods have emerged in recent years. New features with higher quality can be obtained by multi-layer nonlinear transformation of the original features. Scholars at home and abroad have achieved positive results by applying deep learning in the process of speech data to realize PD diagnosis[14][15][16][17][18][19]. However, the performance of deep learning method suffers from the few-shot problem of Parkinson speech data. the deep feature learning directly based on the original features will be better since the dimension of the original features is less than that of the original signal. Besides, as to the existing algorithms, the original features are considered within the deep neural network and training process, which results in poor complementarity between deep features and orignal features, and the unsatisfactory fusion accuracy. In terms of samples, there are low-quality samples due to data collection and other reasons, so it is necessary to reconstruct high-quality samples to improve the classification performance. There is no relevant research report published in the public literatures.

In order to solve these problems, a deep double-side learning ensemble model was proposed for Few-Shot Parkinson Speech Recognition. The main process of the model is briefly described as follows: Firstly, based on the sample sets, the embedded deep stack group sparse auto-encoder is designed to extract deep features based on the original feature, thereby finishing feature reconstruction. Secondly, the original features and deep features are combined and handled with the L1 regularization feature selection algorithm. Then, a deep sample learning algorithm based on iterative mean clustering is designed and deal with the compressed feature set, thereby constructing a deep sample space. Finally, every classifier ies trained based on every sample space, and Bagging ensemble learning mode is used for fusing the classification results.

The main contributions and innovations of this paper is:
- An embedded deep stacked group sparse auto-encoder(EGSAE) is designed for feature learning of PD speech data that integrates the original features of PD speech into the deep network's structure and training. It realizes deep reconstruction of PD speech features, which has improved the complementarity between deep features and original features, thereby improving feature fusion performance.
- A multi-step feature learning mechanism is designed based on EGSAE and L1 regularization feature selection algorithm(L1R&FS), to obtain the better feature representation of the PD speech data.
- A deep sample learning algorithm(DSL) based on iterative mean clustering is constructed for PD speech feature data. The algorithm can generate new PD speech samples by sample transformation, thereby constructing hierarchical sample space, which is helpful to improve the learning performance.
- Based on bagging integrated learning mode(BEM), a deep double-side learning ensemble model(DDSLEM) is constructed by combining EGSAE with DSL, which is helpful to improving the accuracy of few-shot PD speech recognition.

The framework of this paper is briefly described as follows: 1) In the method section, the general idea of the algorithm proposed in this paper is explained, and the main innovative part of the algorithm is introduced, including embedded deep stack group sparse auto-encoder, L1 regularization feature selection, deep sample learning algorithm, etc. 2) In the experimental section, the experimental conditions and data sets are introduced, the effect of parameters on the performance of the algorithm is analyzed, the effectiveness of the algorithm is verified, and the algorithm is compared with the relevant state of art algorithms. 3) In the discussion and conclusion section, the research results, innovation and limitations of this paper are summarized, which lays a foundation and reference value for the subsequent related research.

## 2  Methods

## 2.1 Symbol definitions

Before introducing the algorithm in this paper, the author defines and explains the main symbols involved in this paper as follows:

In this paper, PD speech sample data set is denoted as $X$: $X = \begin{bmatrix} \vec{X_1} \\ \vec{X_2} \\ \vdots \\ \vec{X_M} \end{bmatrix} = \begin{bmatrix} x_{11} & x_{12} & \cdots & x_{1N} \\ x_{21} & x_{22} & \cdots & x_{2N} \\ \vdots & \vdots & \ddots & \vdots \\ x_{M1} & x_{M2} & \cdots & x_{MN} \end{bmatrix}$, sample vector $\vec{X_i} = [x_{i1}, x_{i2}, \cdots, x_{iN}], 1 \leq i \leq M$, devotes the eigenvector corresponding to the $i$-th sample. Class label vector $Y = [y_1, y_2, \cdots, y_M]^T, 1 \leq i \leq M$, where $y_i$ devotes the label value of $i$-th sample, M represents the total number of samples in the data set, and N represents the total number of features in the data set.

In the process of deep feature extraction, the input data at the network input layer is the original data set: $X = [\vec{X_1}, \vec{X_2}, \cdots, \vec{X_M}]^T$. The output data at the hidden layer is eigenvector set $H = [\vec{H_1}, \vec{H_2}, \cdots, \vec{H_M}]^T$. The output data at the output layer is vector set $Z = [\vec{Z_1}, \vec{Z_2}, \cdots, \vec{Z_M}]^T$. The weight matrix between the layers $W = [\vec{W_1}, \vec{W_2}, \cdots, \vec{W_l}]^T$, threshold matrix $B = [\vec{B_1}, \vec{B_2}, \cdots, \vec{B_l}]^T$, where $l$ is the number of hidden layers.

After feature fusion, data set matrix $\hat{X} = \begin{bmatrix} \vec{X'_1} \\ \vec{X'_2} \\ \vdots \\ \vec{X'_M} \end{bmatrix} = \begin{bmatrix} \hat{x}_{11} & \hat{x}_{12} & \cdots & \hat{x}_{1N'} \\ \hat{x}_{21} & \hat{x}_{22} & \cdots & \hat{x}_{2N'} \\ \vdots & \vdots & \ddots & \vdots \\ \hat{x}_{M1} & \hat{x}_{M2} & \cdots & \hat{x}_{MN'} \end{bmatrix}$, feature vectors $\vec{X'_i} = [\hat{x}_{i1}, \hat{x}_{i2}, \cdots, \hat{x}_{iN'}], 1 \leq i \leq M$. The class label vector corresponding to the fused data set remains unchanged: $Y = [y_1, y_2, \cdots, y_M]^T, 1 \leq i \leq M$. Where the total number of the samples in the data set is unchanged, and the total number of features in the sample data set after fusion is $N'$.

The main flow chart of the proposed methodi is shown in Figure 1. Firstly, an embedded deep stack group sparse auto-encoder is designed for deep feature extraction based on PD speech feature data set. Secondly, the deep feature is combined with the original feature to construct the feature set which needs to be optimized. Thirdly, L1 regularization feature selection algorithm is designed for feature compression to form the optimized PD feature set. Then, a deep sample learning algorithm based on iterative mean clustering is designed to construct a deep sample space (hierachical sample space). After that, SVM is trained independently in every layer of sample space .Finally, bagging ensemble mechanism is applied to combine the deep learning algorithm and the deep sample learning algorithm.together, thereby realizing the deep double-side learning ensemble model for PD speech recognition.

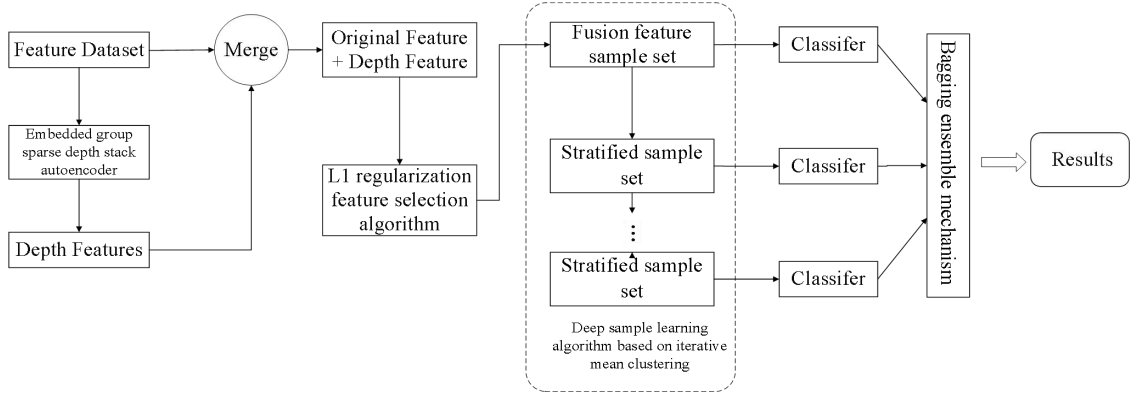

Figure 1 Main flow chart of the proposed algorithm in this thesis

## 2.2 Proposed method - Deep double-side learning ensemble model

### 2.2.1 Embedded deep stacked group sparse auto-encoder(EGSAE)

Deep stack auto-encoder are adopted as the research prototype since it is popular. The deep networks have clear principles, easy implementation, and few parameters. Based on the deep network, group sparse units are added to facilitate the compression of sample dimensions within the neural network. Based on the deep stacked group sparse auto-encoders, the modified embedded unit is designed by introducing the original features into the network structure and training, so as to improve the complementarity between the deep features and the original features. The flow chart of the deep stacked group sparse auto-encoder is shown in Figure 2.

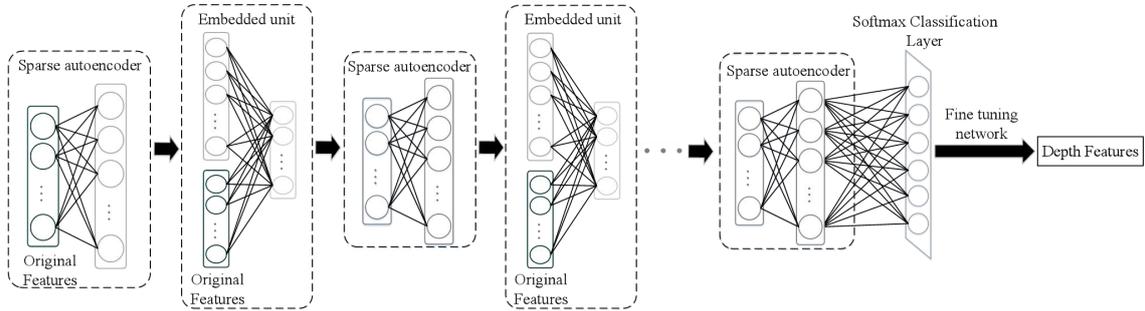

Figure 2 Flow chart of the proposed algorithm

The output data of the hidden layer of the $k$-th sparse encoder is $H^{(k)} = \{h_1^{(k)}, h_2^{(k)}, \cdots, h_N^{(k)}\}$, the number of hidden layer units is $d^{(k)}$. After importing the embedded unit, the matrix $Z^{(k-1)}$ is obtained by connecting the output feature of the previous hidden layer $H^{(k-1)}$ with the original eigenvector $X$. The input data of the $k$-th automatic encoder is

$$L(Z^{(k-1)}) = G^T Z^{(k-1)} \qquad (1)$$

Where the transformation matrix G consists of 0 and 1. To filter low-quality features, the objective function of the embedded unit is:

$$\max_G \ tr(G^T Z^{(k-1)} (Z^{(k-1)})^T G)$$
$$s.t. \ \sum G_{ij} = d \qquad (2)$$

All hidden units are divided into two groups, $\Gamma_1$ and $\Gamma_2$, corresponding to the original features and the output features of the previous layer respectively. The output of hidden layer is expressed as $H = \{H_{\Gamma_1}, H_{\Gamma_2}\}$, group punishment restrictions added to H are shown below.

$$\psi(H) = \sum_{g=1}^{2} \| H_{\Gamma_g} \|_1 \quad (3)$$

In order to add the sparse constraints in the coding process and reduce the dimension of samples effectively, KL divergence is used as the sparse penalty term in this paper. The objective function of the sparse automatic encoder at the $k$-th layer is as follows.

$$\arg\min_{\theta} \frac{1}{N} \sum_{i=1}^{N} \| L(Z^{(k-1)}) - L'(Z^{(k-1)}) \|^2 + \lambda(\|W_{k1}\|_2 + \|W_{k2}\|_2) + \mu(\sum_{j=1}^{d^{(k)}} KL(\rho \| \hat{\rho}_j)) + \sum_{g=1}^{2} \| H_{\Gamma_g}^{(k)} \|_1) \quad (4)$$

Where $W_{k1}$ and $W_{k2}$ are respectively the weight matrix of encoding and decoding. $\lambda$ and $\mu$ are L1 regularizer and sparse penalty term respectively.

Finally, the Softmax layer is connected to the top of the embedded deep stacked group sparse auto-encoder and the entire network is fine tuned in supervised method. The embedded deep stacked group sparse auto-encoder is shown in Table 1.

Table 1 Embedded deep stacked group sparse auto-encoder algorithm

| |
|---|
| **Input:** Original Features |
| **Output:** Deep Features |
| 1: Parameter Setup: $\lambda$, $\mu$, $d^{(k)}$, iterations $l$. |
| 2: pre-traing: |
| 3:   Train the first embedded group sparse autoencoder and extract the encoder output $H^{(1)}$ |
| 4:   for $k = 2, 3, \ldots, K$ |
| 5:     Calculate transformation matrix $G$ |
| 6:     Embed the original feature into the feature $H^{(k-1)}$ |
| 7:     Train the $k$-th layer of the embedded group sparse autoencoder |
| 8:     Extract the output of the $k$-th hidden layer $H^{(k)}$ |
| 9:   end pre-traing |
| 10: Stack the hidden layers and add the Softmax layer to the top |
| 11: Fine tune the entire neural network |
| 12: Output data from the final hidden layer as the deep features |
| 13: end |

**2.2.2 L1 regularization feature selection algorithm(L1R&FS)**

In this paper, L1 regularization constraint is constructed to improve the loss function and realize feature optimization. The optimization function is shown below.

$$\arg\min_{\theta} \sum_{i=1}^{M} (y_i - \sum_{j=1}^{N'} \theta_j \hat{x}_{ij})^2 + \alpha \| \theta \|_1 \quad (5)$$

Where $\alpha \| \theta \|_1$ is the L1 regularizer, $\| \theta \|_1 = \sum_{j=1}^{N} | \theta_j |$, $\theta = [\theta_1, \theta_2, \theta_3, \cdots, \theta_N]$, means the hybrid features including the deep features and original features, and $\theta_j$ represents the regression coefficient of the *j*-th feature; $\alpha$ ($\alpha \geq 0$) is weight coefficient of $L_1$ regularizer.

**2.2.3 Deep sample learning algorithm (DSL) based on iterative mean clustering(IMC)**

In order to analyze the structural information between PD speech samples and fully extract useful information about PD classification, in this paper, an iterative mean clustering algorithm (IMC) is designed to realize deep sample learning (DSL) and build a deep sample space without supervision. The algorithm finds the center of each class by minimizing the distance between the data point and the nearest neighbor center[23], and the center is sent as a new input sample to the next layer of mean clustering, so as to construct the iterative mean clustering algorithm. The main flowchart of the algorithm is shown in Figure 3.

The IMC algorithm is built based on the kernel k-means clustering algorithm (Grigorios&Aristidis,2009)(Li& Hong,2018)(Coates& Ng,2012), which finds each class's center by minimizing the distance between the data points and the nearest neighbor center. The main process of the proposed IMC is described in Figure 2.

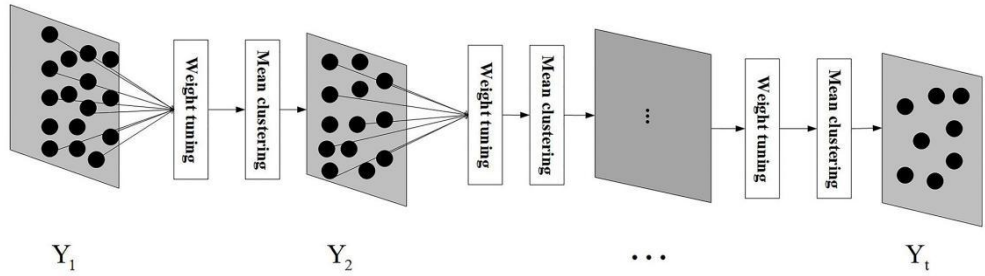

**Figure 3.** A flowchart of the proposed iterative means clustering algorithm (IMC)

The kernel k-means clustering algorithm can be enhanced by the use of a kernel function; where data points are mapped from input space to a higher dimensional feature space through a non-linear transformation, one can extract clusters that are non-linearly separable in input space.

Let us denote clusters by $\pi_k$, and a partitioning of points as $\{\pi_k\}_{k=1}^{K}$. Using the Radial basis function $\phi$, the objective function of kernel k-means is defined as:

$$\min J = \sum_{i=1}^{M}\sum_{k=1}^{K}\overline{\gamma}_{ik}\left\|\phi(\mathrm{x}_i)-\overline{u}_k\right\|$$

$$\text{s.t. } \overline{\gamma}_{ik} \in \{0,1\} \quad \sum_{k=1}^{K}\overline{\gamma}_{ik}=1, i=1,2,3,...,M$$

(6)

where $\overline{u}_k = \dfrac{\sum_{i=1}^{M}\overline{\gamma}_{ik}\phi(\mathrm{x}_i)}{\sum_{i=1}^{M}\overline{\gamma}_{ik}}$, $\overline{u}_k$ is the cluster center.

Suppose the weights of clusters are same, the Euclidean distance from $\phi(\mathrm{x}_i)$ to center $\overline{u}_k$ is given by

$$\left\|\phi(\mathrm{x}_i)-\overline{u}_k\right\|^2 = \phi(\mathrm{x}_i).\phi(\mathrm{x}_i) - \frac{2\sum_{\mathrm{x}_j\in\pi_c}\phi(\mathrm{x}_i).\phi(\mathrm{x}_j)}{|\pi_c|} + \frac{\sum_{\mathrm{x}_j,\mathrm{x}_l\in\pi_c}\phi(\mathrm{x}_j).\phi(\mathrm{x}_l)}{|\pi_c|}$$

(7)

$$\phi(x_i).\phi(x_j)=\phi(x_i.x_j) \qquad (8)$$

In this IMC algorithm, the center point samples clustered with the training data are used as new samples in the subsequent iterations. The algorithm is expressed as follows.

**Table 2** Iterative means clustering (IMC) algorithm

| |
|---|
| ***Iterative means clustering* (IMC) *algorithm*** |
| **Inputs:** training set Y |
| **Output**: the new training dataset and optimal number of sample layers $l$ |
| 1   Initialization: Weight distribution of training samples, $S_1 = (\partial_{11}, \partial_{12},...,\partial_{1m})$; $\partial_{1i} = \frac{1}{m}$, regression error threshold $E$ on training set |
| 2   Define the dictionary size: $D \in R^{k \times M^1}$ ;// $k$ is the number of center points, $M^1$ represents the column dimension |
| 3   For $l = 1:t$ |
| 4      $D\left(\{\pi_k\}_{k=1}^{K}\right)=\arg\min\|\phi(x_i)-\bar{u}_k\|^2$ |
| 5      $\bar{u}_k = \dfrac{\sum_{i=1}^{M}\bar{\gamma}_{ik}\phi(x_i)}{\sum_{i=1}^{M}\bar{\gamma}_{ik}}$ |
| 6      $A^{(K)} = \dfrac{1}{K}\sum_{k=1}^{K}a_k$ // The process of fine tuning. $A^{(K)}$ is the $K_{th}$ row of the dictionary $A \in R^{\theta \times 1}$, $A$ is the label of the output samples, $a_k$ is the age of each instance that belongs to the same cluster. |
| 7      Obtain the new training set $Y_l = [D, A]$ |
| 8      $Y_l \xrightarrow{regress} h_l$ // obtain regression model |
| 9      Calculate the maximum error on training set, $E_l$    Update the weights of the samples, $S_l$ |
| 10      If $E_l \leq E$ |
| 11        save the data set $Y_l$ and the optimal number of sample layers $l$ |
| 12      End |
| 13   End |

**2.2.4 Bagging based Ensemble Model based on Deep Sample Space(BEM)**

As we know, the bagging ensemble learning model[24] is more robust than a sigle model and has higher generalization errors. The prediction of each sub-classifier in the deep sample space are fused by bagging ensemble learning mode. The ensemble learning model is described below.

**Table 3** Bagging based ensemble learning model

**Input**: Speech features $D = \{D_1, D_2, \ldots, D_M\}$, the number of iterations $T$

**Output**: Test Results *Accuracy*

1: Correctarr=Ones(T,1)     // Define the result matrix of SVM classifier
2: For i=1:T
3:     Correctarr(i,1)=Baggingtrainmodel(D)*Correctarr(i,1)
            // Record SVM classification results of each layer of sample sets
4: End For
5: Accuracy=Sum(Correctarr)/T    // Output the final classification results of integration learning
6: End

# 3 Experimental results and analysis

## 3.1 Experimental Conditions

In this paper, the hardware experiment platform's CPU is Intel i5-6200U, and it's memory is 4GB. The algorithm's training part uses the software platform matlab R2018b. The main parameters of the method include: L1 regularization step size is 0.01, relative noise tolerance is 0.01, random noise standard deviation is 0.01; Support Vector Machine (SVM) uses Radial Basis Function kernel (RBF kernel); the number of iterations is 200.

In this paper, the LSVT_voice_rehabilitation data set (Data set 1) and the Sakar data set (Data set 2) are selected as the target set to verify the effectiveness of this algorithm. The two data sets are representative. The first data set is about PD voice based treatment evaluation, and the second data set is about PD voice diagnosis.

**Data set 1-LSVT_voice_rehabilitation data set**: This data set contains information of 14 subjects, 14 of them are Parkinson's patients (6 females, 8 males), each subject has 9 voice samples of different pronunciation tasks (6 Collected before treatment, 3 collected after treatment), and each sample has 309 features.

**Data set 2-Sakar data set: The data set:** created by Sakar et al. contains information of 40 subjects, including 20 Parkinson patients (14 males, 6 females) and 20 normal people (10 males, 10 Women), each subject has 26 speech samples with different pronunciation tasks, and each sample has 26 features. Table 4 summarizes the two data sets.

**Table 4** Summary of Data set

| Data Set | Male/Female | Number of subjects | Samples of each subject | Label 0/1 | Total number of sample | Number of feature |
|---|---|---|---|---|---|---|
| Data Set 1 | 8/6 | 14 | 9 | 84/42 | 126 | 309 |
| Data Set 2 | 24/16 | 40 | 26 | 520/520 | 1040 | 26 |

(label 0 means PD patients, label 1 means normal people)

Based on the two data sets, the Leave One Subject Out (LOSO) is used to do cross-validation. Compared with the hold-out and k-fold cross-validation method, LOSO can avoid the phenomenon that the training sample and the test sample come from the same subject better, so the accuracy of this verification method is closer to the accuracy in the practical application . In addition, the cross-validation method can use of training samples maximally, so for small data set, it can better evaluate the potential of the algorithm, and can compare the accuracy of different algorithms in a better manner.

After testing, the final number of layers in the depth sample space is determined to be 3. Random noise standard deviation is added to enhance L1 regularization's anti-interference ability.

Table 5 Algorithm's parameter settings

| Algorithm | Symbol | Meaning | Parameter Setting |
|---|---|---|---|
| Depth Algorithm | - | encoder layers | 3 |
| | - | number of iterations | 1000 |
| | - | regular coefficient | 0.007 |
| | - | sparse constraint | 4 |
| | - | Sparsity proportion | 0.01 |
| L1 Regular | sigma | random noise deviation | 0.01 |
| | lambda | regularization step | 0.01 |
| | rel_tol | relative target tolerance | 0.01 |

## 3.2 Verification of the proposed Algorithm

In order to verify the effectiveness of the main innovative parts of the proposed algorithm in this paper, the complete version of this proposed algorithm of this paper, and four simplified versions of the proposed algorithm in this paper were realized. The four simplified versions were called -control group 1, control group 2, control group 3 and control group 4. Comparison between them can show the effectiveness of the main innovative parts of the proposed algorithm.

The control group 1 indicates that the algorithm in this paper includes the L1 regularization algorithm and bagging ensemble mode, but not includes IMC; The comparison between the complete version and the control group 1 can verify the effectiveness of deep sample learning algorithm based on IMC. The control group 2 indicates that only the deep features extracted by the EGSAE are used for classification; the comparison between the complete version and the control group 2 is used for verifying the effectiveness of L1R&FS and DSL algorithms. The control group 3 indicates that the original features combined with the deep feature are handled by the L1R&FS algorithm. The comparison between the complete version and the control group 3 is used to verify the effectiveness of the BEM. The control group 4 indicates that only the orignal features are directly used for classification; the comparison between the complete version and the control group 4 is used to verify the effectiveness of the EGSAE and the L1R&FS.

Table 6 shows the average accuracy and the best accuracy of the two data sets using the complete algorithm and the four control group algorithms, respectively. As shown in the table, the accuracy rate of the complete algorithm on data set 1 is 98.1%, which is improved from 95.5% to 98.1% compared with the control group 1; and improved from 82.4% to 98.1 compared with the control group 2; improved from 84.0% to 98.1% compared with the control group 3; improved from 97.8% to 98.1% compared with the control group 4 increased. The accuracy rate of the complete algorithm on data set 2 is 99.1%, which has an improvement of 6.9% compared with control group 1; an improvement of 35.4% compared with the control group 2; an improvement of 32.7% compared with the control group 3 ; an improvement of 0.2% compared with the control group 4. In general, the main innovative parts of the algorithm is effective.

Table 6　Comparison with control groups(average / best)(%)

| Algorithm | Complete Algorithm | Control Group 1 | Control Group 2 | Control Group 3 | Control Group 4 |
|---|---|---|---|---|---|
| Data Set 1 | **98.1/100** | 95.5/99.5 | 82.4/89.5 | 84.0/92.5 | 97.8/100 |
| Data Set 2 | **99.1/100** | 92.2/97.0 | 63.7/72.5 | 66.4/76.0 | 98.8/100 |

## 3.3 Comparison with the relevant algorithms

In order to further verify the effectiveness and advantages of the proposed algorithm, it is compared with existing related algorithm. For Data set 1, currently only the relevant algorithm is publicly reported in the literature [25]. The literature uses the same data set as this paper; the involved feature selection algorithm is LOGO (fit locally and think globally), and the cross-validation method used is Leave-One-Subject-Out (LOSO), and the SVM is used for classification. In short, the experimental conditions of the two algorithms are similar, so the comparison results can verify the performance of two algorithms fairly. The classification accuracy of this paper is compared with the classification accuracy of the mentioned literature in Figure 4. The blue line is the accuracy rate of the proposed algorithm, the green line is the classification accuracy rate of the literature's algorithm, the abscissa is the number of features used, and the ordinate is the classification accuracy. It can be seen from the figure, that the classification accuracy rate concentrates at 95%~100% by using the proposed algorithm, and the classification accuracy rate concentrates at 85%~90% when using the iterature's algorithm. There is a clear difference between the two algorithms, indicating that the proposed method can effectively improve the classification accuracy.

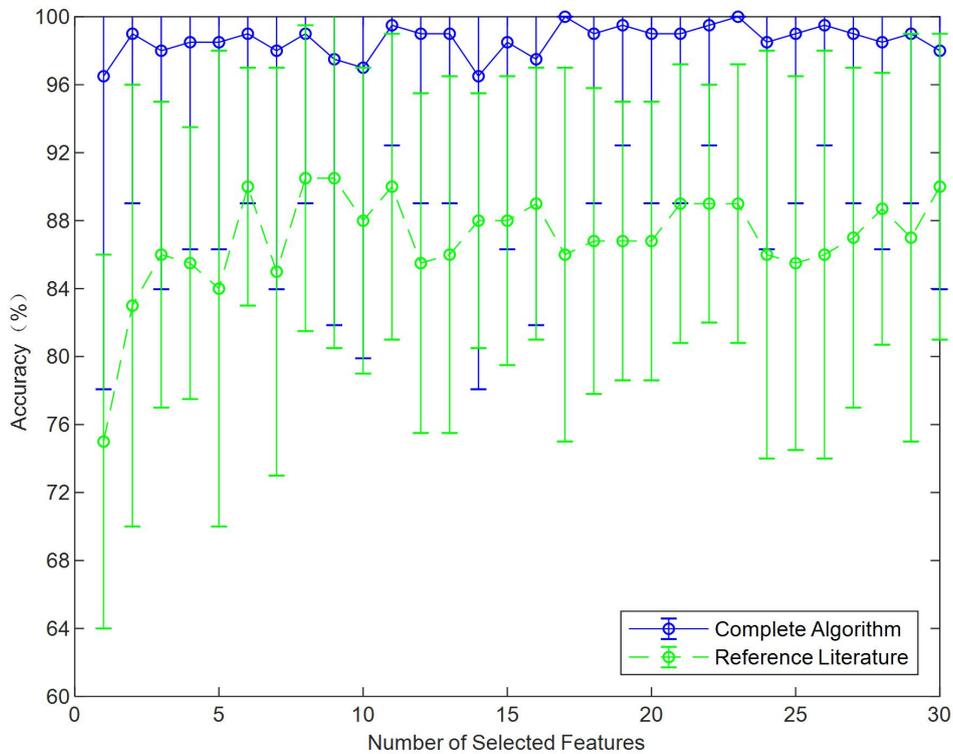

Figure 4 Comparison of the proposed algorithm and the releveant algorithm on data set 1

Table 7 is the average accuracy and the best accuracy of this proposed algorithm and the literature's algorithm. The highest accuracy rate of the literature's algorithm is 90%, and the highest accuracy rate of this proposed algorithm is 8% better than the literature's algorithm. In addition, the mean value of the proposed algorithm is as high as 98.4%, and the mean value of specificity is as high as 99.7%. In addition, the low standard deviation of the proposed algorithm shows its high stability and great practical potential. The possible reason for this difference is that there is no fusion of deep features and original features with L1 regularization in that literature. In addition, there is no deep sample learning part in this compared algorithm, which may also be one of the reasons for its low accuracy.

**Table 7** Comparison of Literature and This Thesis(%)

| Algorithm | Accuracy | Sensitivity | Specificity |
|---|---|---|---|
| | Mean / Best | Mean / Standard | Mean / Standard |
| Complete Algorithm | **98.4/100** | **98.4/0.009** | **99.7/0.016** |
| Control Group 1 | 96.4/99.5 | 96.4/0.014 | 97.9/0.023 |
| Comparison Literature | 87/90 | — | — |

For Data set 2, there are many related research literatures. In order to compare the performance of the algorithm more fairly, Table 8 directly lists the accuracy rates reported in the relevant literature. It can be found from the table that, in addition to the proposed algorithm in this paper, there are only three algorithms with a classification accuracy rate of more than 90%. Classification accuracy rate Zhang's [29] LSVM+MSVM+RSVM+CART+KNN+LDA+NB algorithm can reach 94.17%, however, this algorithm has a very low sensitivity, only 50%, and the specificity is 94.92%. Khan's [36] Evolutionary Neural Network Ensembles algorithm has a classification accuracy of 90%, sensitivity of 93%, and specificity of 97%. The classification accuracy of this thesis is improved by 9.6%, sensitivity by 6.3%, and specificity increased by 2.8%. Classification accuracy, sensitivity, and specificity of Ali's [20] LDA-NN-GA algorithm are all up to 95%. The classification accuracy of the remaining algorithms in the table are less than 90%, which is far lower from accuracy rate of the proposed algorithm. In conclusion, it can be clearly seen that the classification accuracy, sensitivity and specificity of the proposed method are the best compared with other methods in the table, so it shows that the proposed method is effective in terms of classification accuracy.

It should be noted that the methods of some documents here is based on hold-out and k-fold for cross-validation. The two methods tend to cause different samples come from the same subject, that is, the training sample and the test sample possibly come from same subject, so the accuracy rate is often higher than the accuracy rate under LOSO (false high). Therefore, if the accuracy of the two algorithms is the same, the algorithm based on LOSO verification is often better. Even so, the accuracy rate of the proposed algorithm under LOSO in this paper is still the highest, indicating that the proposed method is significantly effective.

Table 8 Classification Results of Different Methods(Data set 2)

| Literature | Algorithm | Accuracy (%) | Sensitivity (%) | Specificity (%) |
|---|---|---|---|---|
| Sakar et al. [1] | KNN+SVM | 55 (LOSO CV) | 60 | 50 |
| Canturk and Karabiber [26] | 4 Feature Selection Methods+ 6 Classifiers | 57.5 (LOSO CV) | 54.28 | 80 |
| Eskidere et al.[27] | Random Subspace Classifier Ensemble | 74.17 (10-fold CV) | — | — |
| Behroozi and Sami[28] | Multiple classifier framework | 87.50 (A-MCFS) | 90.00 | 85.00 |
| Zhang et al.[29] | MENN+RF with MENN | 81.5 (LOSO CV) | 92.50 | 70.50 |
| Benba et al.[30] | HFCC+SVM | 87.5 (LOSO CV) | 90.00 | 85.00 |
| Li et al.[31] | Hybrid feature learning+SVM | 82.50 (LOSO CV) | 85.00 | 80.00 |
| Vadovsk and Paralic[32] | C4.5+C5.0+RF+CART | 66.5 (4-fold CV) | — | — |
| Zhang [33] | LSVM+MSVM+RSVM+CART+KNN+LDA+NB | 94.17 (Holdout) | 50.00 | 94.92 |
| Benba et al[34] | MFCC+SVM | 82.5 | 80.0 | 85.0 |

| | | (LOSO CV) | | |
|---|---|---|---|---|
| Kraipeerapun and Amornsamanku [35] | Stacking+CMTNN | 75 (10-fold CV) | — | — |
| Khan et al.[36] | Evolutionary Neural Network Ensembles | 90 (10-fold CV) | 93.00 | 97.00 |
| Ali et al.[20] | LDA-NN-GA | 95 (LOSO CV) | 95 | 95 |
| - | DBN | 54.6 (LOSO CV) | 52.4 | 56.8 |
| - | CNN | 60.0 (LOSO CV) | 63.0 | 57.0 |
| - | DBN+SVM | 50.5 (LOSO CV) | 53.0 | 48.0 |
| - | DBN+SVM(TL) | 55.5 (LOSO CV) | 60.0 | 51.0 |
| - | Autoencoder+SVM | 67.5 (LOSO CV) | 65.0 | 70.0 |
| - | Autoencoder+SVM(TL) | 72.5 (LOSO CV) | 75.0 | 70.0 |
| | Propose algorithm | **99.6** (LOSO CV) | **99.3** | **99.8** |

## 4  Conclusion and discussion

Speech feature learning is the core of PD diagnosis, therapeutic effect evaluation and other application methods, and has the characteristics of few-shot learning. Therefore, the general method is to extract the speech features and then to conduct feature learning. Existing feature learning mainly includes feature selection and feature extraction. The former is not conducive to forming new features; the latter can not generate high-quality new features through multi-layer transformation. Although deep feature learning has the advantage of high-efficiency automatic feature learning ability, which can form new features through multi-layer transformation, it is mainly aimed at the original speech signal and relies heavily on large sample size. Therefore, it will be better to conduct deep learning on the original speech features and combine the deep features and the original features. Until now, there are no similar study about the PD speech recognition. Besides, as to the existing methods about fusion of deep features and original features,the deep feature extraction process does not consider traditional features, resulting in poor complementarity between deep features and original features, and high redundancy after the fusion of the two types of the features. In order to further solve the few-shot learning problem, it is necessary to consider the feature reduction and ensemble learning.

To solve these problems above, a deep double-side learning ensemble model was proposed for PD speech recognition. The major parts of the proposed algorithm is briefly described as follows: Firstly, based on the existing PD feature datasets, an embedded deep stacked group sparse autoencoder (EGSAE) is designed for dealing with the PD feature datasets, thereby reconstructing the original features and extracting deep features. The EGSAE considered the original features within network and during training process. Then, the original features and deep features are merged, and are handled by L1 regularized feature selection algorithm (L1R&FS ) which reduces the feature dimension and improves the generalization ability. Next, a deep sample learning algorithm (DSL) based on iterative mean clustering (IMC) is designed for reconstructing and augmenting the data samples. New samples are formed by clustering, and deep sample space is constructed to fully mine the structural information between samples. Finally, a classifier is trained based on each sample space, and bagging ensemble learning mode (BEM) is used to fuse the results of each sub-classifier. In order to fully verify the performance of this proposed algorithm, two representative public PD speech datasets are used about PD speech diagnosis and efficacy evaluation, and several representative algorithms are compared. The experimental results show that the mean accuracy of the proposed algorithm can reach 98.4% and 99.6% on the two data sets.

The main innovations and contributions of this work include: 1) For the PD speech feature data, an embedded group sparse depth stack autoencoder is constructed, which considers the original features into the deep network structure and training, realizes the deep reconstruction of the PD speech features, improves the complementarity between the deep features and the original features, and helps to improve the feature fusion performance. 2) A multi-step feature learning mechanism of embedded group sparse depth stack autoencoder + L1 regularization feature compression is constructed, which improves the feature fusion ability. 3) For the PD speech feature data, a deep sample learning algorithm based on iterative mean clustering is constructed, and a hierarchical sample space is constructed, which realizes the deep reconstruction of the PD speech samples and obtains new PD speech samples, which is helpful to improve the sample learning performance. 4) Based on bagging ensemble learning mode, a deep double-side learning ensemble model is constructed by combining embedded deep stacked group sparse autoencoder and deep sample learning algorithm, which is helpful to improve the accuracy of speech recognition for small sample PD. In a word, the major contribution and innovation is to propose a deep double-side learning ensemble model to better balance hybrid feature representation and dimensions for PD speech recognition.

Although some literatures on PD speech recognition have involved feature learning research, including feature selection methods (Relief, LASSO, etc), feature transformation methods (PCA, LDA,etc) and deep feature learning method. However according to the authors' knowledge, these studies seldom involve deep learning research on PD speech features directly to balance the deep feature learning and few-shot learning. Besides, there are no public literature reports that integrate the original speech features into the structure and training of deep network. What is more remarkable is that there is no publicly available literature reporting the use of iterative mean clustering for PD deep sample learning to reconstruct and augment the PD speech samples. This paper not only proposes a novel concept - deep double-side learning, but also applies it to PD speech recognition, which preliminarily shows its effectiveness and advantages.

Although the proposed algorithm is novel and effective, there are some works to do in the near future. Firstly, more datasets and deep neural networks needs to be consdiered to further verify the proposed method; secondly, it is necessary to study how to better reconstruct the samples compared the iterative means clustering algorithm.